\documentclass[twoside,11pt]{article}

\usepackage{blindtext}

\usepackage[preprint]{jmlr2e}

\usepackage{epsfig} 
\usepackage{times} 
\usepackage{amsmath} 
\usepackage{amssymb}  
\usepackage{amsfonts}
\usepackage{mathptmx} 
\usepackage{mathtools}
\usepackage{tabularx}
\usepackage{epstopdf}
\usepackage{textcomp}
\usepackage{float}
\usepackage{epic,color}
\usepackage{mathrsfs}
\usepackage{dsfont,MnSymbol}
\usepackage{algorithm}
\usepackage{algorithmic}
\usepackage{multirow} 
\usepackage[T1]{fontenc}
\usepackage{booktabs}

\usepackage{tikz}
\usetikzlibrary{arrows, positioning, calc}
\usetikzlibrary{arrows.meta}
\usetikzlibrary{fit}
\usepackage{pgfplots}
\usepgfplotslibrary{groupplots}

\def\Re{\mathbb{R}}

\def\Sec#1{Sec.~\ref{#1}}
\def\Fig#1{Fig.~\ref{#1}}

\def\notes#1{\marginpar{\tiny #1}\typeout{Notes!
Notes!
Notes!
}}
\renewcommand{\notes}[1]{\typeout{notes!}}
\def\FRAC#1#2#3{\genfrac{}{}{}{#1}{#2}{#3}}
\def\half{{\mathchoice{\FRAC{1}{1}{2}}%
{\FRAC{2}{1}{2}}%
{\FRAC{3}{1}{2}}%
{\FRAC{4}{1}{2}}}}

\def\Re{\field{R}}

\def\Sec#1{Sec.~\ref{#1}}

\def\transpose{{\hbox{\rm\tiny T}}}

\def\clP{{\cal P}}
\def\clZ{{\cal Z}}

\def\Sec#1{Sec~\ref{#1}}

\def\E{{\sf E}}

\def\Fig#1{Fig.~\ref{#1}}
\def\Sec#1{Sec.~\ref{#1}}

\def\clZ{{\cal Z}}

\newtheorem{assumption}{Assumption}
\newtheorem{hypothesis}{Hypothesis}
\newtheorem{objective}{Objective}

\def\beq{\begin{eqnarray}} 
\def\bc{\begin{center}} 
\def\be{\begin{enumerate}}
\def\bi{\begin{itemize}} 
\def\bs{\begin{small}}
\def\bS{\begin{slide}}
\def\ec{\end{center}} 
\def\ee{\end{enumerate}}
\def\ei{\end{itemize}}
\def\es{\end{small}}
\def\eS{\end{slide}}
\def\eeq{\end{eqnarray}}

\newcommand{\newP}[1]{\medskip\noindent{\bf #1:}}

\def\Re{\mathbb{R}}
\def\E{{\sf E}}

\def\clY{{\cal Y}}

\def\Sec#1{Sec.~\ref{#1}}
\def\Thm#1{Thm.~\ref{#1}}
\def\Prop#1{Prop.~\ref{#1}}

\def\clD{{\cal D}}

\def\clP{{\cal P}}
\def\clZ{{\cal Z}}

\renewcommand{\Re}{\mathbb{R}}

\def\FRAC#1#2#3{\genfrac{}{}{}{#1}{#2}{#3}}

\def\clD{{\cal D}}

\def\clN{{\cal N}}

\def\clP{{\cal P}}

\def\clU{{\cal U}}

\def\clY{{\cal Y}}
\def\clZ{{\cal Z}}
\def\E{{\sf E}}
\def\bS{\mathbb{S}}
\def\bO{\mathbb{O}}
\def\sJ{{\sf J}}

\def\ones{{\sf 1}}
\def\sP{{\sf P}}

\def\tp{{\hbox{\rm\tiny T}}}

\def\dvar{\operatorname{var}}

\def\kl{{\sf D}}

\def\opt{{\text{\rm (opt)}}}
\def\det{{\text{\rm (det)}}}
\def\xfer{{\text{\rm (xfer)}}}

\usepackage{lastpage}
\jmlrheading{23}{2025}{1-\pageref{LastPage}}{}{}{25-0000}{H.-S. Chang and P. G. Mehta}

\ShortHeadings{Transformer and its probabilistic modeling}{Chang and Mehta}
\firstpageno{1}

\begin{document}

\title{What can we learn from signals and systems in a transformer?
  Insights for probabilistic modeling and inference architecture}

\author{\name Heng-Sheng Chang \email hschang@illinois.edu \\
       \addr Coordinated Science Laboratory\\
       University of Illinois Urbana-Champaign\\
       Urbana, IL 61801, USA
       \AND
       \name Prashant G. Mehta \email mehtapg@illinois.edu \\
       \addr Coordinated Science Laboratory\\
       University of Illinois Urbana-Champaign\\
       Urbana, IL 61801, USA}

\editor{My editor}

\maketitle

\begin{abstract}
  In the 1940s, Wiener introduced a linear predictor, where the future
  prediction is computed by linearly combining the past
  data. A transformer generalizes this idea: it is a nonlinear predictor where the next-token
  prediction is computed by nonlinearly combining the past tokens.  In
  this essay, we present a probabilistic model that interprets
  transformer signals as surrogates of conditional measures, and layer
  operations as fixed-point updates.  An explicit form of the
  fixed-point update is described for the special case when the
  probabilistic model is a hidden Markov model (HMM).  In part, this paper is in
  an attempt to bridge the classical nonlinear filtering theory with modern
  inference architectures. 
\end{abstract}


\section{Introduction}\label{sec:intro}

A recent book by Stephen Wolfram is titled
\begin{quote}
{\em What Is ChatGPT Doing
  … and Why Does It Work?}
\end{quote}
Our essay is written as a response to the first of Wolfram's two
questions. Before doing so, we must clarify our interpretation of the
meaning of these questions, while acknowledging that our understanding
may differ from Wolfram's own intentions\footnote{We also asked ChatGPT
to clarify the meaning of Wolfram's questions. ChatGPT's answer is
reproduced verbatim in the Appendix~\ref{appdx:chatgpt_Wolfram}.}.

\paragraph{1. What Is ChatGPT Doing?} A transformer is an algorithmic
solution to a certain problem---that of predicting the next token.
Our goal in this essay, which we view as a partial answer to Wolfram's
question, is to carefully define the token prediction problem, discuss
its key difficulties, and speculate on the mathematical
insights that a transformer solution reveals.

\paragraph{2. Why Does it Work?}  This is a more difficult question that
we do not attempt to address in this essay. In our view, the question
highlights a central mystery: how does merely solving for next-token
prediction problem lead to the coherent generation of words,
sentences, and even entire paragraphs?

\medskip

Our focus in this essay is on the problem related to 
Wolfram's first question.  While we discuss some of the details of a
transformer solution, the essay is {\em not} intended as an
explanation of the attention mechanism.  In fact, part of our
motivation comes from a concern that many explanations of attention
focus narrowly on mechanistic details, while leaving the underlying
problem unexamined.

Our goal in this essay is two-fold:
\begin{enumerate}
\item Formulation of the mathematical problem that a transformer
  solves.  This is referred to as {\em probabilistic modeling} and
  formalized via a set of assumptions and hypotheses. 
\item Enunciation of modeling and algorithmic insights that the
  transformer solution reveals. This is referred to as the
  transformer-inspired {\em inference architecture}. 
\end{enumerate}

\subsection{The Transformer and its historical antecedents}

\begin{figure}[t!]
\centering
\begin{tikzpicture}[
  transformer/.style={draw, thick, minimum width=5cm, minimum height=1.5cm},
  inputarrow/.style={-Latex, thick},
  outputarrow/.style={-Latex, thick}
]

\node[transformer] (transformer) at (0,0) {Transformer};

\foreach \i/\x in {1/-2, 2/-1, T/2} {
  \node at (\x, -2) (z\i) {$z_{\i}$};
  \draw[inputarrow] (z\i) -- (\x, -0.75);
}

\node at (0.5, -2) (dots) {$\ldots$};

\node at (2, 2) {$p_{T}$};
\draw[outputarrow] (2,0.75) -- (2,1.8);

\node at (-3, -2) {(input)};
\node at (4, 2) {(output)};

\end{tikzpicture}
\caption{Function of a transformer: Input to a transformer is a
  sequence of tokens $(z_1,z_2,\hdots,z_T)\in\bO^T$ and the output is
  the conditional probability vector $p_T$.}
\label{fig:tranformer}
\end{figure}
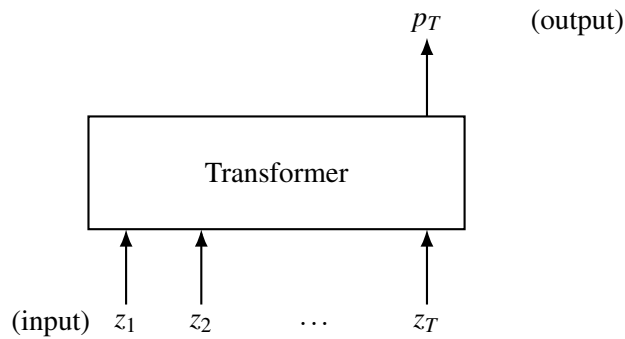

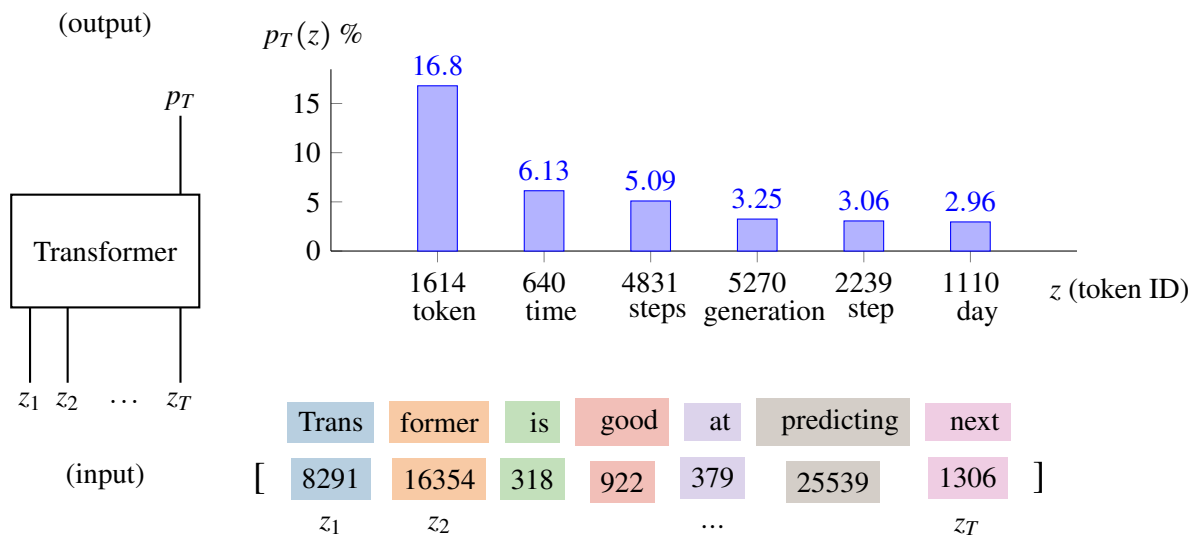
\begin{figure}[t!]
\centering
\begin{tikzpicture}[
    axis/.style={thick, ->},
    bar/.style={draw, fill=gray!20},
    tick/.style={},
    word/.style={minimum height=0.5cm, draw=none, inner sep=4pt, anchor=north, text centered},
    idbox/.style={minimum height=0.4cm, draw=none, inner sep=4pt, anchor=north, text centered, text=black},
    transformer/.style={draw, thick, minimum width=2.5cm, minimum height=1.5cm},
    inputarrow/.style={thick},
    outputarrow/.style={thick}
  ]

  \begin{scope}[shift={(-3, 2)}]

    \node[transformer] (transformer) at (0,0) {Transformer};

    \foreach \i/\x in {1/-1, 2/-0.5, T/1} {
      \node at (\x, -2) (z\i) {$z_{\i}$};
      \draw[inputarrow] (z\i) -- (\x, -0.75);
    }

    \node at (0.25, -2) (dots) {$\ldots$};

    \node at (1, 2) {$p_{T}$};
    \draw[outputarrow] (1,0.75) -- (1,1.8);

    \node at (0, -3) {(input)};
    \node at (0, 3) {(output)};
  \end{scope}

  \begin{scope}[shift={(0,2)}] 
    \begin{axis}[
      ybar, 
      ymin=0,
      height=4cm, 
      width=11.5cm, 
      enlarge x limits={0.1, 0.2}, 
      xtick=data, 
      xticklabels={
        \shortstack{1614\\ ~~token},
        \shortstack{640\\ ~~time},
        \shortstack{4831\\ ~~steps},
        \shortstack{5270\\ ~~generation},
        \shortstack{2239\\ ~~step},
        \shortstack{1110\\ ~~day}
      },
      ylabel={$p_T(z)~\%$},
      xlabel={$z$ (token ID)},
      nodes near coords, 
      nodes near coords align={vertical},
      bar width=15pt,
      axis x line*=bottom,
      axis y line*=left,
      ylabel style={at={(axis description cs:0.1,1.05)},anchor=south,rotate=-90},
      xlabel style={at={(axis description cs:0.95,0)},anchor=west}
    ]
    \addplot coordinates {
      (1, 16.8)
      (2, 6.13)
      (3, 5.09)
      (4, 3.25)
      (5, 3.06)
      (6, 2.96)
    };
    \end{axis}
  \end{scope}

  \definecolor{c1}{RGB}{184, 207, 224}
  \definecolor{c2}{RGB}{248, 202, 165}
  \definecolor{c3}{RGB}{195, 224, 187}
  \definecolor{c4}{RGB}{242, 191, 184}
  \definecolor{c5}{RGB}{220, 211, 235}
  \definecolor{c6}{RGB}{212, 206, 200}
  \definecolor{c7}{RGB}{239, 205, 227}
  
  \newcommand{\toksep}{0.5em}

  \node[word, fill=c1] (w1) at (0, 0) {Trans};
  \node[word, fill=c2, right=\toksep of w1] (w2) {former};
  \node[word, fill=c3, right=\toksep of w2] (w3) {~~is};
  \node[word, fill=c4, right=\toksep of w3] (w4) {~~good};
  \node[word, fill=c5, right=\toksep of w4] (w5) {~~at};
  \node[word, fill=c6, right=\toksep of w5] (w6) {~~predicting};
  \node[word, fill=c7, right=\toksep of w6] (w7) {~~next};
  
  \node[idbox, fill=c1, below=0.2cm of w1] (id1) {8291};
  \node[idbox, fill=c2, below=0.2cm of w2] (id2) {16354};
  \node[idbox, fill=c3, below=0.2cm of w3] (id3) {318};
  \node[idbox, fill=c4, below=0.2cm of w4] (id4) {922};
  \node[idbox, fill=c5, below=0.2cm of w5] (id5) {379};
  \node[idbox, fill=c6, below=0.2cm of w6] (id6) {25539};
  \node[idbox, fill=c7, below=0.2cm of w7] (id7) {1306}; 
  
  \node[anchor=east, font=\Large] at ($(id1.west) + (-0.2,0)$) {[};
  \node[anchor=west, font=\Large] at ($(id7.east) + (0.2,0)$) {]};
  
  \node[anchor=south, below=1.4cm of w1] at (w1.north) {$z_1$};
  \node[anchor=south, below=1.4cm of w2] at (w2.north) {$z_2$};
  \node[anchor=south, below=1.5cm of w5] at (w5.north) {$\cdots$};
  \node[anchor=south, below=1.4cm of w7] at (w7.north) {$z_T$};
  
\end{tikzpicture}
\caption{Empirical results from a numerical example illustrating the inference process in a
transformer: (bottom) the language prompt ``Transformer is good at
predicting next'' is tokenized into a sequence
$(z_1,z_2,\ldots,z_T)\in\bO^T$; (top) the transformer outputs the
conditional distribution $p_T(z)$ for the next token, with the six most
likely tokens shown. The results are obtained using the open-source
nanoGPT model.}
\label{fig:inference}
\end{figure}

 Let $\bO=\{0,1,2,\hdots,m\}$ denote a finite set called the vocabulary.
An element of $\bO$ is referred to as a token.  A
sequence of $T$ tokens is an $\bO^T$-valued random vector, denoted by
$\{Z_1,Z_2,\hdots,Z_T\}$.  A decoder-only transformer is an algorithm to
compute the conditional probability of the next token (see Fig.~\ref{fig:tranformer}):
\[
p_T(z):=\sP(Z_{T+1}=z \mid Z_1,Z_{2},\hdots,Z_T),\quad z\in \bO.
\]
During inference with a well-trained transformer, the conditional
probability is often sparse---that is, only a small subset
of tokens has non-negligible probability (see Fig.~\ref{fig:inference}). This sparsity is useful 
for efficient sampling in generative AI
applications~\cite{wolfram2023chatgpt}.

There are two distinguishing features of the decoder-only transformer
architecture:
\begin{enumerate}
\item Even though only the conditional probability at the terminal time
  $t=T$ is of interest, conditional probabilities are also computed
  for intermediate times,
  \[
p_t(z):=\sP(Z_{t+1}=z \mid Z_1,Z_2,\hdots,Z_t),\quad z\in \bO,\quad t=1,2,\hdots,T.
\]
\item In all cases, the conditional probability of the next token is
  represented as a causal, nonlinear function of the past tokens,
  implemented through a procedure known as causal masking. In this
  essay, we refer to such a representation as a {\em nonlinear predictor}.  
\end{enumerate}
The second item is in contrast to a
recurrent neural network (RNN) architecture, where a hidden state is
stored and recursively updated~\cite{graves2013generating,dai2019transformer}.  

\newP{Linear predictor} A transformer architecture is reminiscent of the classical Wiener
filter. Recall that a Wiener filter computes the conditional
expectation of a Gaussian process, also denoted (with
slight abuse of notation) as $[Z_1,Z_2,\hdots,Z_T]$, in the following
causal form:
\begin{equation}
\text{(linear predictor)}\qquad \E(Z_{T+1} \mid Z_1,Z_{2},\hdots,Z_T) = \text{(constant)} + \sum_{t=0}^{T-1} u_t^\transpose Z_{t+1}. \label{eq:linear_predictor}
\end{equation}
The right-hand side is an example of a linear predictor where
$u:=\{u_t\in\Re^{m\times m}:0\leq t\leq T-1\}$ are deterministic
weights, to be
designed or learned.  The Wiener
filtering theory is concerned with the synthesis of the optimal weights that
yield the conditional expectation~\cite[Ch.~7]{kailath2000linear}.

\medskip

A generalization of the linear predictor
representation~\eqref{eq:linear_predictor}, now applicable to the 
$\bO$-valued stochastic processes, is given in the following
sub-section.

\subsection{Representation formula for nonlinear predictor}
Consider a vector-valued mapping $e:\bO\to \Re^m$ as
follows:
\[
e(1) = \begin{bmatrix} 1 \\ 0 \\ \vdots \\ 0 \end{bmatrix}_{m\times 1},\quad e(2)
= \begin{bmatrix} 0 \\ 1\\ \vdots\\ 0 \end{bmatrix}_{m\times 1}, \quad \hdots
\quad e(m)
= \begin{bmatrix} 0 \\ 0 \\ \vdots \\ 1 \end{bmatrix}_{m\times 1}, \quad e(0) =
-e(1) -e(2) - \hdots -e(m).
\]
(Recall here that the cardinality $|\bO| = m+1$.)  

\begin{example}[m=1]
  Suppose the observations are binary-valued, i.e., $\bO =
  \{0,1\}$.  Then
  \[
e(1) = 1,\quad e(0) = -1.
\]
\end{example}
In the nonlinear settings, the space of
admissible weights (control inputs) is as follows:
\[\clU  :=\{U:\Omega\times \{0,1,2,\hdots,T-1\} \to \Re^m, \;\;
  U_t\in\clZ_t, \;0\leq t\leq T-1\},
\]
where $\clZ_t:=\sigma(\{Z_s:1\leq s\leq t\})$ (sigma-algebra of
observations up to time $t$) and $\clZ_0:=\{\phi,\Omega\}$.  The definition is useful to obtain the
following key representation (compare with~\eqref{eq:linear_predictor}):
\begin{equation}\label{eq:nonlin_predictor_rep}
\text{(nonlinear predictor)}\qquad \sP(Z_{T+1}=z \mid Z_1,Z_2,\hdots,Z_T) = \text{(const.)} - \sum_{t=0}^{T-1} U_t^\transpose
e(Z_{t+1}),\;\; \sP\text{-a.s.},\quad z\in\bO
\end{equation}
where $U=\{U_0,U_1,\hdots,U_{T-1}\}\in\clU$.  The right-hand side of~\eqref{eq:nonlin_predictor_rep} is referred to as a
{\em nonlinear predictor} and $U$ is referred to as the control
input.  The well-posedness result for the same is
given in the following.

\begin{proposition}\label{prop:existence_nonlin_predictor_rep} 
For each $z\in \bO$, there exists $U\in\clU$
such that~\eqref{eq:nonlin_predictor_rep} holds. Suppose 
$\sP(z_1,z_2,\hdots,z_T)>0$ for all $(z_1,z_2,\hdots,z_T)\in \bO^T$
then the process $U$ is unique. 
\end{proposition}
\begin{proof}
See Appendix~\ref{appdx:existence}.  
\end{proof}

\begin{remark}[Comparison between linear and nonlinear predictors]
While the weights $u$ in a linear predictor~\eqref{eq:linear_predictor} are
deterministic, the weights in a nonlinear predictor~\eqref{eq:nonlin_predictor_rep} are random---i.e., $U_t$ is allowed to depend
upon past observations $\{Z_1,Z_2,\hdots,Z_t\}$ for each $0\leq t\leq
T-1$.  The key point is that either of these representations describe a
causal transformation of the data.  See Table~\ref{tab:lin_non_lin_predictor}.
\end{remark}
\begin{table}[t!]
\centering
\renewcommand{\arraystretch}{1.5}
\begin{tabular}{|l|c|l|c|}
\hline
\textbf{Algorithm} & \textbf{Object computed} &
                                                      \textbf{Representation}
  & \textbf{Model Assmp.}\\ \hline
Linear predictor &
$\E(Z_{T+1} \mid Z_1, \ldots, Z_T)$ &
$= \text{(const.)} + \sum_{t=0}^{T-1} u_t^\transpose Z_{t+1}$ & $Z_t$ is Gaussian\\ \hline
Nonlin.~predictor &
$\sP(Z_{T+1} \mid Z_1, \ldots, Z_T)$ &
$= \text{(const.)} + \sum_{t=0}^{T-1} U_t^\transpose e(Z_{t+1})$ & $Z_t$ is
                                                             $\bO$-valued
  \\ \hline
\end{tabular}
\caption{Comparison of linear and nonlinear predictors. In the linear
  case, $u$ is a deterministic process. In the nonlinear case, $U$ is
  a $\clZ$-adapted stochastic process.}
\label{tab:lin_non_lin_predictor}
\end{table}

\begin{remark}[Connection to transformer]\label{remark:connection_to_transformer}
An aim of this work is to interpret the transformer as an
efficient algorithm for computing the weights $U$
in~\eqref{eq:nonlin_predictor_rep}. 
The attention mechanism provides a convenient parameterization for
this computation, linking the abstract nonlinear predictor to the
concrete architecture of a transformer.  
\end{remark}

\subsection{Outline}

The outline of the remainder of this essay is as follows: In~\Sec{sec:xfer_prob}, a self-contained description of the transformer
is presented together with insights for the probabilistic modeling.
The modeling is used to motivate a fixed-point
architecture for inference (formalized as Objective~\ref{obj:goal}).  For
the special case of an HMM, the architecture is described in
\Sec{sec:HMM} (based on our prior work in~\citep{chang2025dual}).  The paper closes with some directions for future
research in \Sec{sec:conc}. All the proof appear in the
Appendix~\ref{appdx}.

\section{Transformer and its probabilistic modeling}\label{sec:xfer_prob}

As noted in \Sec{sec:intro}, a transformer is an algorithm to
compute the conditional probability $p_T\in \clP(\bO)$ (the set of
probability vectors on $\bO$) for any given sequence of input
$(z_1,z_2,\hdots,z_T)\in \bO^T$. Therefore,   
$p_T$ may be regarded as a nonlinear function,
\[
p_T : \bO^T \to \clP(\bO).  
\]

\newP{Complexity of the prediction problem} 
A key difficulty is the size of the domain: the cardinality of \(
\bO^T \) is \( (m+1)^T \), where note \( m+1 = |\bO|
\). Table~\ref{tab:transformer_params} tabulates typical values of $m$
and $T$ used in open source models together with an estimate of
$(m+1)^T$.  These are very large numbers.  It is therefore quite remarkable that a transformer is able to
learn a (good enough) representation of this function!

\begin{table}[h!]
\centering
\begin{tabular}{|l|c|c|c|}
\hline
\textbf{Model}       & \textbf{Vocabulary size} \( (m +1)\) & \textbf{Context length} \( T \) & \textbf{Estimate of \( (m+1)^T \)} \\ \hline
GPT-2                & 50,257                  & 1,024                   & \( \approx 10^{3084} \)       \\ \hline
GPT-3                & 50,257                  & 2,048                   & \( \approx 10^{6168} \)       \\ \hline
BERT (Base)          & 30,000                  & 512                     & \( \approx 10^{2457} \)       \\ \hline
LLaMA 2              & 32,000                  & 4,096                   & \( \approx 10^{17{,}221} \)   \\ \hline
\end{tabular}
\caption{Comparison of transformer model parameters in terms of the estimated
  number of distinct input sequences. For most models, this number far
  exceeds the number of atoms (\( \approx 10^{68} \)) in the Milky Way galaxy.} 
\label{tab:transformer_params}
\end{table}

\newP{How to reduce complexity?}
That a transformer is able to learn a representation of $p_T$ is
because of the correlations that exist in human language.  We have
already defined the observation process 
\[
\text{(observation process)} \qquad Z := [Z_1,Z_2,\hdots,Z_T,Z_{T+1}],
\]
which is a stochastic process takings values in $\bO$.  
A time honored approach to model the correlations is through the
introduction of a hidden (latent) process, 
\[
\text{(hidden state process)} \qquad X:=[X_0,X_1,\hdots,X_{T-1},X_T].
\]
The law of the joint process is expressed as 
\[
\sP(X,Z) = \sP(Z\mid X) \sP(X),
\]
so that modeling the joint behavior amounts to specifying \(\sP (Z
\mid X)\) and \(\sP(X)\).  

\newP{Key questions, assumptions, and hypotheses}
Concerning this model, the following questions are of interest:
\begin{enumerate}
\item What is the meaning of the state $X$?
\item How are the internal signals in a transformer related to $X$?
\end{enumerate}
Towards constructing a probabilistic model for the transformer, we
make the following assumptions concerning the state $X$ and the 
factorization of the model $\sP(Z\mid X)$ (a justification is given in
the following sub-section):

\begin{assumption}\label{assmp:state-space}
The state process $X$ takes values in the state-space
$\bS=\{1,2,\hdots,d\}$.  
\end{assumption}

\begin{assumption}\label{assmp:conditional-independence}
The following conditional independence property holds:
  \begin{equation}
\sP(Z_{t+1}=z
\mid X_t=x,Z_1,Z_2,\hdots, Z_t) = \sP(Z_{t+1}=z
\mid X_t=x)=:C(x,z), \quad x\in\bS, \quad z\in\bO,\quad t\in\mathbb{T}.     \label{eq:conditional_indep}
\end{equation}
\end{assumption}


Denote
\[
\pi_t(x) := \sP(X_t=x \mid Z_1,Z_2,\hdots,Z_t),\quad x\in\bS,\quad t=1,2,\hdots,T.
\]
The definition is useful to describe the following formula, which
relates $\pi_t$ to $p_t$. 
\begin{proposition}\label{prop:formula_for_ptz}
Suppose Assumptions~\ref{assmp:state-space}
and~\ref{assmp:conditional-independence} hold.  Then 
\begin{align}\label{eq:formula_pt_pit}
p_t(z) = \sum_{x\in \bS} \pi_t(x) C(x,z), \quad z\in\bO,\quad t=1,2,\hdots,T.
\end{align}
\end{proposition}
\begin{proof}
  The proof is elementary using the law of total probability,
  \begin{align*}
p_t(z) = \sP(Z_{t+1}=z\mid Z_1,Z_2,\hdots, Z_t) &= \sum_{x\in \bS} \sP(Z_{t+1}=z
\mid X_t=x,Z_1,Z_2,\hdots, Z_t) \; \sP(X_{t}=x \mid Z_1,Z_2,\hdots,
                                                  Z_t) \\
  &= \sum_{x\in \bS} \sP(Z_{t+1}=z
    \mid X_t=x) \; \pi_t(x) = \sum_{x\in \bS} C(x,z) \pi_t(x),\quad z\in\bO
\end{align*}
\end{proof}

It then remains to specify a model for $\sP(X)$. A natural causal
decomposition is given by, 
\begin{align*}
  \sP(X) &= \sP(X_0,X_1,\hdots,X_{T-1},X_T) \\
  &= \sP(X_0) \; \sP(X_1 \mid X_0) \; \sP(X_2 \mid X_1,X_0) \; \cdots \;  \sP(X_T \mid X_{T-1},\hdots,X_1,X_0).
\end{align*}

\begin{remark}\label{rem:HMM}\normalfont
The latent process is introduced because modeling $\sP(X)$ is simpler
than modeling $\sP(Z)$.  One possibility is to assume that $X$ is a Markov process, i.e.,
  \[
\sP(X_{t+1} \mid X_t,\hdots,X_0) = \sP(X_{t+1} \mid X_t), \quad 0\leq
t\leq T-1.
\]
This leads to a modeling of the joint process
$(X,Z)$ as an HMM.  It is {\em not} our intent to advocate a first-order HMM as a model for
$(X,Z)$. The example is presented merely to illustrate one instance of
a more general causal factorization of $\sP(X)$. An HMM serves as a useful
starting point.  
\end{remark}

The key hypothesis of this work is as follows:

\begin{hypothesis} \label{hyp:pi}
The operations in a transformer are designed to
compute a \textbf{surrogate} of the conditional measure
$\pi:=\{\pi_t(x):x\in\bS,\;t=1,2,\hdots,T\}$.
\end{hypothesis}

The remainder of this section is divided into two subsections:
\begin{enumerate}
\item In~\Sec{sec:io}, the input and the output signals in a transformer
  are described. The aim is to obtain some insights that are useful for probabilistic
  modeling.  In particular, 
\begin{enumerate}
  \item Justify 
  Assumptions~\ref{assmp:state-space}
  and~\ref{assmp:conditional-independence}; and
  \item Describe a transformer surrogate
    of the conditional measure $\pi$ and provide justification for the
    Hypothesis~\ref{hyp:pi}.
\end{enumerate} 
\item In~\Sec{sec:attention}, the key mathematical operations 
  from the input to the output are described. The aim is to motivate a
  certain fixed-point representation for the solution being computed
  by the transformer.    
\end{enumerate}
The transformer descriptions are based
upon~\cite{phuong2022formal}~\cite[Ch.,~10]{jm3}
and~\cite{raschka2024build}.

\subsection{Transformer signals: Insights for probabilistic modeling}\label{sec:io}

\newP{Input and output signals and their mathematical abstraction}
In a transformer, as a first step, a token is ``embedded'' using a $d\times m$ embedding
matrix denoted in this essay by $C^\xfer$:
\[
z \mapsto C^\xfer (:,z) \in \Re^d,\quad z\in\bO.
\]
Using the embedding matrix,
\[
\begin{bmatrix} z_1 & \hdots & z_t & \hdots &
  z_T \end{bmatrix}_{1\times T}  \mapsto \begin{bmatrix} C^\xfer (:,z_1) & \hdots & C^\xfer (:,z_t) & \hdots & C^\xfer (:,z_T) \end{bmatrix}_{d\times T}.
\]
The input $C^\xfer (:,z_t)$ is augmented with the so-called positional
encoding as follows:
\[
\sigma^{(0)}_t(x) := C^\xfer(x,z_t) + W_p(x,t),\quad x\in\bS, \quad t=1,2,\hdots,T,
\]
where $W_p \in \Re^{d\times T}$ is referred to as the positional encoding matrix.  In
the original transformer paper~\cite{vaswani2017attention}, the rows of $W_p$ are defined
according to the sinusoidal-positional-encoding:
\begin{align*}
  W_p(2i-1,t) = \sin(\ell_{\text{max}}^{-\frac{2i}{d}} t),\quad 
  W_p(2i,t)  = \cos(\ell_{\text{max}}^{-\frac{2i}{d}} t), \quad
  i=1,2,\hdots,\frac{d}{2},\quad t=1,2,\hdots,T,
\end{align*}
where $\ell_{\text{max}} = 10,000$.  The factor
$\ell_{\text{max}}^{-\frac{2i}{d}}$ determines the frequency of
oscillations, ensuring a wide range of scales, as $i$ varies from
$1,2,\hdots,\frac{d}{2}$.  The positional encoding is the {\em only}
mechanism through which information about the position (time) $t$ is introduced
into the transformer. 
Other types of positional encoding are also possible~\cite{dufter2022position}.

From an input-output perspective, a decoder-only transformer implements a causal nonlinear
transformation that transforms a $d\times T$ matrix at the
input into a $d\times T$ matrix at the output,
\[
[\sigma_1^{(0)}, \sigma_2^{(0)},\hdots,
  \sigma_T^{(0)} ]_{d\times T} \mapsto [\sigma_1^{(L)}, \sigma_2^{(L)},\hdots,\sigma_T^{(L)} ]_{d\times T}.
\]
Then the conditional probability of the
next token is computed as follows: 
\[
p_{t}(z)=\sP(Z_{t+1}=z \mid Z_1=z_1,\hdots,Z_t=z_t) = \frac{e^{((\sigma_t^{(L)})^\tp
    C^\xfer)(z)}}{\sum_{z'\in\bO} e^{((\sigma_t^{(L)})^\tp C^\xfer)(z')}},\quad
t=1,2,\hdots,T,\quad z\in\bO.
\]
The operation on the right-hand side is referred to as softmax.

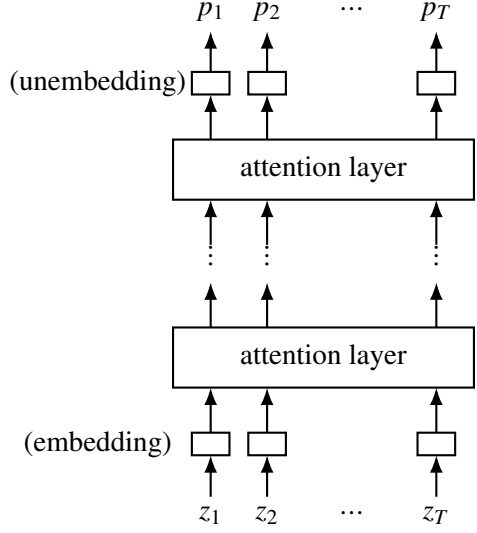
\begin{figure}[t!]
  \centering

      \begin{tikzpicture}[
  layerBox/.style={draw, thick, minimum width=4cm, minimum height=0.8cm},
  embeddingBox/.style={draw, thick, minimum width=0.5cm, minimum height=0.3cm},
  arrow/.style={-Latex, thick}
]
\def\timeXPositions{-1.5, -0.75, 1.5}
  \def\timeLabels{1, 2, T}
  \def\initLayerY{-1.25}
  \def\finalLayerY{1.25}
  \def\arrowLength{0.6}
  \def\layerHeight{0.8}
  \def\boxHeight{0.3}
  \def\beddingX{-3}
  \def\embeddingY{\initLayerY - \layerHeight/2 - \arrowLength - \boxHeight/2}
  \def\unembeddingY{\finalLayerY + \layerHeight/2 + \arrowLength + \boxHeight/2}
  \def\inputY{\embeddingY - \boxHeight/2 - \arrowLength - 0.2}
  \def\outputY{\unembeddingY + \boxHeight/2 + \arrowLength + 0.2}

  \newcommand{\atAllTimePositions}[1]{%
    \foreach [count=\i] \x in \timeXPositions {
      \foreach [count=\j] \label in \timeLabels {
        \ifnum\i=\j #1 \fi
      }
    }
  }

  \newcommand{\verticalArrowsAt}[2]{%
    \foreach \x in \timeXPositions {
      \draw[arrow] (\x, #1) -- (\x, #2);
    }
  }

  \foreach \name/\y in {initial/\initLayerY, final/\finalLayerY} {
    \node[layerBox] (layer\name) at (0,\y) {attention layer};
    \verticalArrowsAt{\y - \layerHeight/2 - \arrowLength}{\y - \layerHeight/2}
    \verticalArrowsAt{\y + \layerHeight/2}{\y + \layerHeight/2 + \arrowLength}
  }

  \foreach \embeddingName/\y in {embedding/\embeddingY, unembedding/\unembeddingY} {
    \node at (\beddingX, \y) {(\embeddingName)};
    \foreach \x in \timeXPositions {
      \node[embeddingBox] at (\x, \y) {};
    }
  }

  \atAllTimePositions{
    \node at (\x, \inputY) (z\label) {$z_{\label}$};
    \draw[arrow] (z\label) -- (\x, \embeddingY - \boxHeight/2);
  }
  
  \atAllTimePositions{
    \node at (\x, \outputY) (p\label) {$p_{\label}$};
    \draw[arrow] (\x, \unembeddingY + \boxHeight/2) -- (p\label);
  }

  \foreach \x in \timeXPositions {
    \node at (\x, 0.1) {$\vdots$};
  }
  \foreach [count=\i] \x in \timeXPositions {
    \ifnum\i=2 \xdef\secondPos{\x} \fi
    \ifnum\i=3 \xdef\thirdPos{\x} \fi
  }
  \pgfmathsetmacro{\midpoint}{(\secondPos + \thirdPos) / 2}
  \node at (\midpoint, \inputY) {$\cdots$};
  \node at (\midpoint, \outputY) {$\cdots$};

\end{tikzpicture}
    
  \caption{Transformer architecture from the input
    $(z_1,z_2,\hdots,z_T)$ to the output $(p_1,p_2,\hdots,p_T)$.}
  \label{fig:transformer}
  
\end{figure}

Internally, a transformer is arranged in $L$ layers as follows (see \Fig{fig:transformer}):
\begin{align*}
&\text{(input)} \quad [z_1,z_2,\hdots,z_T]_{1\times T} \mapsto
                 [\sigma_1^{(0)} , \sigma_2^{(0)} ,\hdots, \sigma_T^{(0)} ]_{d\times T} \qquad
                 \text{(embedding + positional-encoding)}\\[5pt]
&\text{(first layer)} \quad   [\sigma_1^{(0)} , \sigma_2^{(0)} ,\hdots, \sigma_T^{(0)} ]_{d\times T} \mapsto [\sigma_1^{(1)},\sigma_2^{(1)},\hdots,\sigma_T^{(1)}]_{d\times T}\\[5pt]
  &\text{(intermediate layer)} \quad  [\sigma_1^{(\ell)},\sigma_2^{(\ell)},\hdots,\sigma_T^{(\ell)}]_{d\times T}  \mapsto  [\sigma_1^{(\ell+1)},\sigma_2^{(\ell+1)},\hdots,\sigma_T^{(\ell+1)}]_{d\times T},\quad \ell=1,2,\hdots,L-1
  \\[5pt]
  &\text{(output)} \quad
    [\sigma_1^{(L)},\sigma_2^{(L)},\hdots,\sigma_T^{(L)}]_{d\times
    T} \mapsto [p_1,p_2,\hdots,p_T]_{m\times T} \qquad \text{(un-embedding)}
\end{align*}

\newP{Insights for probabilistic modeling}
Our task is to attach a meaning to a single layer operation in a
transformer. 
Specifically,
\begin{enumerate}
\item What is the meaning of the input
  $\sigma^{\ell}:=[\sigma_1^{(\ell)},\sigma_2^{(\ell)},\hdots,\sigma_T^{(\ell)}]$?
\item What is the meaning of the output
  $\sigma^{\ell+1}:=[\sigma_1^{(\ell+1)},\sigma_2^{(\ell+1)},\hdots,\sigma_T^{(\ell+1)}]$?
\item What is the meaning of the transformer operation from the
  input $\sigma^{\ell}$ to the output $\sigma^{\ell+1}$?
\end{enumerate}

Recall that the output of the last layer is used to compute the conditional
probability of interest:
\begin{subequations}
\begin{equation}\label{eq:unembedding_xfer}
   \text{(transformer)}\qquad \ln p_t(z) =  \sum_{x\in\bS} \sigma_t^{(L)}(x) C^\xfer(x,z) + \text{(constant)},\quad t=1,2,\hdots,T, \quad z\in\bO. 
\end{equation}
This is compared with the formula~\eqref{eq:formula_pt_pit} for conditional measure in
\Prop{prop:formula_for_ptz}, as repeated below,
\begin{equation}\label{eq:unembedding_pm}
  \text{(probabilistic model)}\qquad p_t(z) = \sum_{x\in \bS} \pi_t(x) C(x,z),\quad t=1,2,\hdots,T, \quad z\in\bO.
\end{equation}
\end{subequations}
The above two equations provide the \textbf{key correspondence} that
is used to motivate the following:
\begin{enumerate}
\item Interpretation of $\sigma_t^{(L)}=\{\sigma_t^{(L)}(x):x\in\bS\}$ as a
  surrogate of the conditional measure
  $\pi_t=\{\pi_t(x):x\in\bS\}$, for $t=1,2,\hdots,T$, where $\bS$ has
  an interpretation of 
  the state-space.  
\item Justification of the conditional independence
property~\eqref{eq:conditional_indep} described as part of 
Assumption~\ref{assmp:conditional-independence}.  Indeed, using the
definitions of $p_t$ and $\pi_t$, an application of the total probability lemma shows that
\begin{align*}
p_t(z) = \sP(Z_{t+1}=z\mid Z_1,Z_2,\hdots, Z_t) = \sum_{x\in \bS}
                                                  \pi_t(x) \; \sP(Z_{t+1}=z
\mid & X_t=x,Z_1,Z_2,\hdots, Z_t),\\ &\qquad z\in\bO,\quad t=1,2,\hdots,T.
\end{align*}
Based on the correspondence between $\sigma_t$ and $\pi_t$,
formula~\eqref{eq:unembedding_xfer} in the transformer provides a
justification for assuming the 
conditional independence property~\eqref{eq:conditional_indep} which
in turn gives the formula~\eqref{eq:unembedding_pm} in terms of the
conditional measure.  
\end{enumerate}
The correspondence is tabulated in the
Table~\ref{tab:correspondence}. While $\pi_t$ is an element of the space of
  probability measures on $\bS$, its surrogate
  $\sigma_t^{(L)}$ is an element of the space of signed
  measures on $\bS$.  The two spaces are denoted by $\clP(\bS)$ and
  ${\cal M}(\bS)$, respectively.  Then
  \[\pi=[\pi_1,\pi_2,\hdots,\pi_T]
  \in \clP(\bS)^T,\quad 
  \sigma^{(L)} =[\sigma_1^{(L)},\sigma_2^{(L)},\hdots,\sigma_T^{(L)}] \in {\cal
    M}(\bS)^T.
  \]

\begin{table}[h]
\centering
\begin{tabular}{|p{3.5cm}|p{5.5cm}|p{5.5cm}|}
\hline
\textbf{Concept} & \textbf{Probabilistic Model} & \textbf{Surrogates in a Transformer} \\
\hline
State space & \( \bS := \{1,2,\ldots,d\} \) & Dimension $d$ of embedding vector \\
\hline
  Observation space & \( \bO := \{0,1,2,\ldots,m\} \) & Vocabulary \\
  \hline
Observation process & \( Z= [Z_1, Z_1, \ldots, Z_T],\ Z_t \in \bO \) &
                                                                    Prompt
  \\
\hline
Latent process & \( X = [X_0, X_1, \ldots, X_T],\ X_t \in \bS \) &
                                                                   Internal
                                                                   latent
                                                                   state
  in language\\
\hline
Observation model & \( C(x,z)=\sP(Z_{t+1} = z \mid X_t = x) \) & Embedding column \( C^\xfer(x, z) \) \\
\hline
Conditional measure & \( \pi_t(\cdot) = \sP(X_t = \cdot \mid Z_1, \ldots, Z_t) \) & Transformer layer output \( \sigma_t^{(L)}\) \\
\hline
\end{tabular}
\caption{Probabilistic model and its surrogates in a transformer.}
\label{tab:correspondence}
\end{table}

\subsection{Transformer operations: Insights for the solution
  architecture} \label{sec:attention}


Because each layer in a transformer is structurally identical, for the
purposes of describing the mapping, it suffices to consider a generic layer whose input
and output are denoted by,
\[
\sigma := [\sigma_1, \sigma_2,\hdots, \sigma_T]_{d\times T} \in  {\cal M}(\bS)^T,\quad \sigma^+:=
[\sigma^+_1, \sigma^+_2,\hdots, \sigma^+_T]_{d\times T} \in {\cal M}(\bS)^T,
\]
respectively, and the layer mapping is denoted by
  \[
  {\cal N}^\xfer:{\cal M}(\bS)^T \mapsto {\cal M}(\bS)^T\quad
  \text{s.t.}\quad \sigma^+ = {\cal N}^\xfer(\sigma).
\]
For the first layer, the input is given by the embedding plus
positional encoding.  For each subsequent layer, the input is defined
by the output of the preceding layer.

Operationally, a layer is comprised of multiple attention heads.  A single attention head is an input-output map of the form
  \[
  [\sigma_1, \sigma_2,\hdots, \sigma_T]_{d\times T} \mapsto
  [o_1^h,o_2^h,\hdots,o_T^h]_{d_V \times T}.
  \]
  where the output vector $o_t^h \in\Re^{d_V}$ whose 
  dimension $d_V = \frac{1}{n_{\text{head}}} d$, where $n_{\text{head}}$
  has the meaning of the number of heads. 
  The operations in a single attention head are defined by three matrices
  \[
  W_V^h \in \Re^{d_V \times d}, \;\;W_Q^h \in \Re^{d_K \times d}, \;\;W_K^h \in \Re^{d_K \times d},
  \]
  where typically $d_K=d_V$. The matrices are constant (fixed) during
  the inference phase of the transformer operation.  

The mathematical operation defining the {\em causal self-attention} is as follows:
  \begin{equation}\label{eq:transf_output}
  o_t^h = W_V^h \left( \sum_{s=1}^t \alpha(s;t,h) \sigma_s\right),\quad t=1,2,\hdots, T,
  \end{equation}
  where $\{\alpha(s;t,h):s=1,2,\hdots,t\}$ is referred to as the
  attention weight vector.  For each fixed $(t,h)\in \{1,2,\hdots,T\}\times\{1,2,\hdots,
  {n_{\text{head}}}\}$, it is
  a probability vector.  That is,
\[
  \alpha(s;t,h)\geq 0,\quad s=1,2,\hdots,t,\quad \text{and}\quad 
  \sum_{s=1}^t \alpha(s;t,h)= 1.
  \]

  The attention weights are computed as follows:  For each fixed $1\leq t\leq T$, 
  define \begin{align*}
  \text{(query)} \qquad q_t &= W_Q^h \sigma_t, \\
           \text{(key)} \qquad k_s &= W_K^h \sigma_s,\quad s=1,2,\hdots, t.
\end{align*}
Then
\[
\text{(attention weight)} \qquad \alpha(s;t,h) = \frac{\exp({\frac{q_t^\transpose k_s}{\sqrt{d_K}}})}{\sum_{\tau=1}^t \exp({\frac{q_t^\transpose k_\tau}{\sqrt{d_K}}})},\quad s=1,2,\hdots, t.
\]
In the terminology adopted in this essay, the dependence of the
weights upon the data (which varies between sample paths of
observations), makes attention a nonlinear predictor.  If the weights
were deterministic (same for all sample paths), attention will be an
example of linear predictor (see Remark~\ref{remark:connection_to_transformer}).

From~\eqref{eq:transf_output}, the
layer output is obtained as 
\[
\sigma^+_t = W_O \; \text{concat}(o_t^1, \hdots, o_t^{n_{\text{head}}}),\quad t=1,2,\hdots,T,
\]
where $W_O \in \Re^{d\times d}$.

  An issue that arises is that the output may `blow up' during the
  training phase of a transformer.  For this reason, additional miscellaneous
  operations are implemented.  All of these operations are carried out
  independently for each fixed $t$.  Because these additional operations are not
  especially pertinent to probabilistic modeling,
  these are described in the Appendix~\ref{sec:xfer}.

\newP{Insights for probabilistic modeling}  Define
\[
L^h :=  W_O \; \text{concat}(0,\hdots,0,W_V^h,0,\hdots,0),\quad h=1,2,\hdots,n_{\text{head}}.
\]
For each fixed $h$, $L^h$ is a $d\times d$ matrix.  Fix
$f\in\Re^d$ and define,
\[
y_s^{(t)} := \sum_{h=1}^{n_{\text{head}}} \alpha(s;t,h) ((L^h)^\transpose f),\quad s=1,2,\hdots,t.
\]
Then the
mapping ${\cal N}:\sigma \mapsto   \sigma^+$ is
succinctly expressed as
\begin{equation}\label{eq:rep_attention_simplified}
\sigma_t^+(f) = \sum_{s=1}^t \sigma_s(y_s^{(t)}),\quad
t=1,2,\hdots,T,\quad f\in\Re^d.
\end{equation}
(here $\sigma_t^+(f) := (\sigma_t^+)^\tp f$ and $\sigma_s(y_s^{(t)})
:=\sigma_s^\tp (y_s^{(t)})$).

\subsection{Numerical results for the example}

\begin{figure}[h]
\centering
\begin{tikzpicture}

  \begin{scope}[shift={(0, 3)}]
    \begin{axis}[
      ymode=log,
      ymin=0.1,
      height=4cm,
      width=13cm,
      ylabel={$\overline{\kl}(\ell)$},
      xlabel={layer $\ell$},
      enlargelimits=0.15,
      xtick=data,
      grid 
    ]
    \addplot[mark=*, mark options={color=black, mark size=2pt}, line width=2pt] coordinates {
      (0, 10.765)
      (1, 4.9817)
      (2, 4.4272)
      (3, 4.1994)
      (4, 4.0273)
      (5, 3.9758)
      (6, 3.864)
      (7, 4.0679)
      (8, 3.9612)
      (9, 3.7107)
      (10, 3.5897)
      (11, 3.0761)
      (12, 2.3588)
      (13, 1.8852)
      (14, 1.438)
      (15, 1.1029)
      (16, 0.89517)
      (17, 0.66642)
      (18, 0.48842)
      (19, 0.27544)
    };
    \end{axis}
  \end{scope}
  \begin{scope}
    \begin{groupplot}[
      group style={group size=1 by 5, vertical sep=0.5cm},
      ybar,
      ymin=0,
      height=3cm,
      width=13cm,
      enlarge x limits={0.2, auto},
      xtick=data,
      axis x line*=bottom,
      axis y line*=left,
      ylabel style={at={(axis description cs:0.1,1.05)},anchor=south,rotate=-90},
      legend style={at={(1.02,1)},anchor=north west, draw=none, fill=none},
    ]

    \nextgroupplot[xticklabels={,,,,,}, ylabel={$p_T(z)~\%$}]
      \addplot[fill=purple] coordinates {
        (1, 16.8)
        (2, 6.13)
        (3, 5.09)
        (4, 3.25)
        (5, 3.06)
        (6, 2.96)
      };
      \legend{layer 20}



    \nextgroupplot[xticklabels={,,,,,}]
      \addplot[fill=orange] coordinates {
        (1, 15.25)
        (2, 0.46)
        (3, 0.79)
        (4, 1.40)
        (5, 2.68)
        (6, 0.63)
      };
      \legend{layer 5}

    \nextgroupplot[xlabel={token ID},
                   xlabel style={at={(axis description cs:1.05,0)},anchor=west},
                   xticklabels={
                     \shortstack{1614\\~~token},
                     \shortstack{640\\~~time},
                     \shortstack{4831\\~~steps},
                     \shortstack{5270\\~~generation},
                     \shortstack{2239\\~~step},
                     \shortstack{1110\\~~day}
                   }]
      \addplot[fill=white] coordinates {
        (1, 19.08)
        (2, 0.17)
        (3, 0.05)
        (4, 0.72)
        (5, 0.73)
        (6, 0.26)
      };
      \legend{layer 1}

    \end{groupplot}
  \end{scope}

\end{tikzpicture}
\caption{(top) $\overline{\kl}(\ell)$ as a function of the layer index
  $\ell$ and (bottom) conditional probability estimate for top six
  tokens computed from the output of first, fifth, and final
  (twentieth) layers.  The numerical results are computed with the
  open-source nanoGPT model using the prompt ``Transformer is good at
  predicting next'' (see Fig.~\ref{fig:inference}).}
\label{fig:layer_output}
\end{figure}
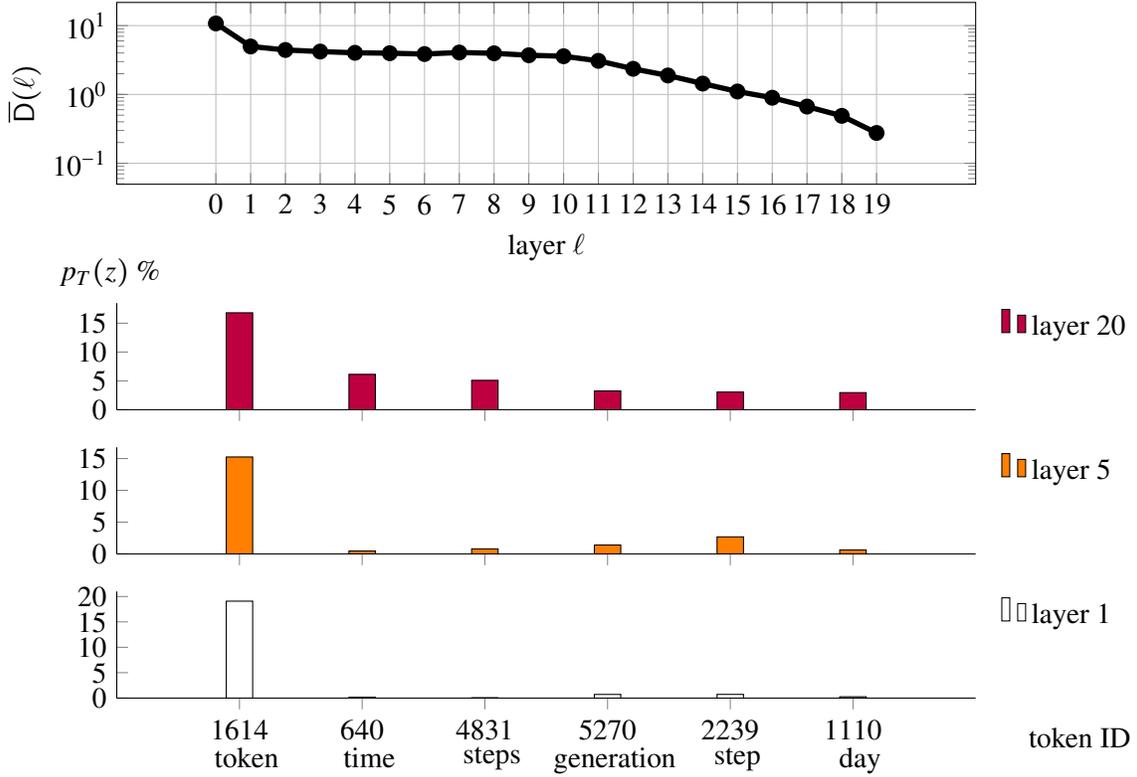

In this sub-section, we revisit the 
example first introduced as part of Fig.~\ref{fig:inference} in \Sec{sec:intro}. 
Even though the output of the last layer is used to compute the
conditional probability of interest, one could do the same with the
output of any intermediate layer.  For this purpose, denote
  \[
p_{t}^{(\ell)}(z):= \frac{e^{((\sigma_t^{(\ell)})^\tp
    C^\xfer)(z)}}{\sum_{z'\in\bO} e^{((\sigma_t^{(\ell)})^\tp C^\xfer)(z')}},\quad z\in\bO, \quad
t=1,2,\hdots,T,\quad \ell=0,1,2,\hdots, L-1.
\]
Consider the following metric to compare predictions of the
intermediate layer with those of the final layer:
\[
\overline{\kl}(\ell) := \frac{1}{T} \sum_{t=1}^{T}\left(\sum_{z\in\bO} p_t^{(L)} (z) \ln \frac{p_t^{(L)} (z)
}{p_t^{(\ell)} (z)}\right),\quad \ell=0,1,2,\hdots,L-1.
\]
For the prompt described in Fig.~\ref{fig:inference}, a numerical plot
of 
$\{\overline{\kl}(\ell):\ell=0,1,,2,\hdots, L-1\}$ is depicted in
Fig.~\ref{fig:layer_output}. It is
seen that the successive layer output provides a monotonically better
approximation.  It is also remarkable that the output of the first
layer already assigns largest probability to ``token''.

\newP{Insights for probabilistic modeling} These numerical results suggest the following:
\begin{enumerate}
\item The layer output $\sigma_t^{(\ell)}\in{\cal
    M}(\bS)$, for 
  $\ell=1,2,\hdots,L-1$, also has the meaning of an (approximate) surrogate of the conditional measure
  $\pi_t\in \clP(\bS)$.
\item The layer mapping $\sigma^{(\ell)}\mapsto
  \sigma^{(\ell+1)}$ is an algorithm to iteratively compute (improve)
  $\sigma^{(L)}$ starting from the initialization $\sigma^{(0)}$.
\end{enumerate}

All of this motivates the following hypothesis regarding the layer operation:

\begin{hypothesis}\label{hyp:FPE}
The surrogate $\sigma^{(L)}$ is a fixed-point of
$\clN^\xfer$.  
\end{hypothesis}

\subsection{Objective: Fixed-point architecture for inference}

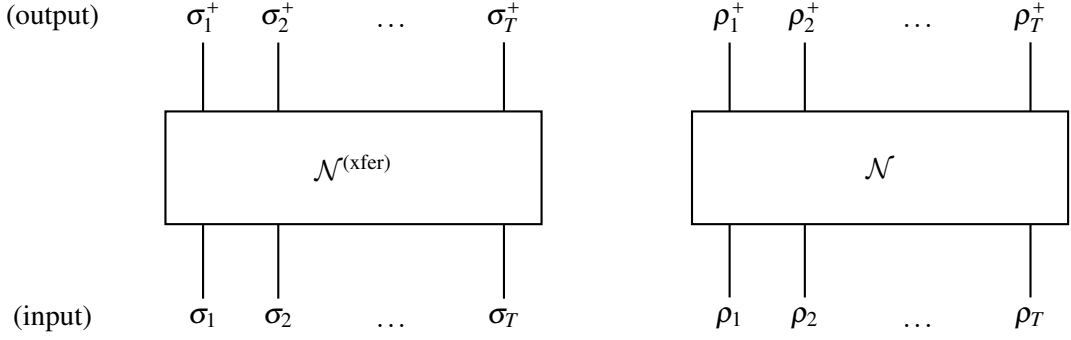
\begin{figure}[t!]
\centering
\begin{tikzpicture}[
  transformer/.style={draw, thick, minimum width=5cm, minimum height=1.5cm},
  inputarrow/.style={thick},
  outputarrow/.style={thick}
]
\begin{scope}[shift={(-7, 0)}]
  \node[transformer] (transformer) at (0,0) {$\clN^\xfer$};

  \foreach \i/\x in {1/-2, 2/-1, T/2} {
    \node at (\x, -2) (z\i) {$\sigma_\i$};
    \draw[inputarrow] (z\i) -- (\x, -0.75);
  }

  \node at (0.5, -2.1) (dots) {$\ldots$};

  \foreach \i/\x in {1/-2, 2/-1, T/2} {
    \node at (\x, 2) (z+\i) {$\sigma_\i^{+}$};
    \draw[outputarrow] (z+\i) -- (\x, 0.75);
  }

  \node at (0.5, 1.9) (dots) {$\ldots$};

  \node at (-4, -2) {(input)};
  \node at (-4, 2) {(output)};
\end{scope}

\begin{scope}
  \node[transformer] (transformer) at (0,0) {$\clN$};

  \foreach \i/\x in {1/-2, 2/-1, T/2} {
    \node at (\x, -2) (z\i) {$\rho_\i$};
    \draw[inputarrow] (z\i) -- (\x, -0.75);
  }

  \node at (0.5, -2.1) (dots) {$\ldots$};

  \foreach \i/\x in {1/-2, 2/-1, T/2} {
    \node at (\x, 2) (z+\i) {$\rho_\i^{+}$};
    \draw[outputarrow] (z+\i) -- (\x, 0.75);
  }

  \node at (0.5, 1.9) (dots) {$\ldots$};
\end{scope}

\end{tikzpicture}
\caption{(left) A transformer layer mapping $\clN^\xfer:\sigma\mapsto
  \sigma^+$ is given by~\eqref{eq:rep_attention_simplified}.
 (right) Research objective~\ref{obj:goal} is to design the mapping 
$\clN:\rho\mapsto\rho^+$ such that the conditional measure
$\pi^{(z)}$ is a fixed-point of $\clN$.}
\label{fig:transformer_objective}
\end{figure}

Because the transformer layer output is interpretted as a surrogate of the conditional
measure, our objective is to develop analogous
representation (similar to~\eqref{eq:rep_attention_simplified}) for
$\pi$.  For this purpose, consider any fixed sample path of observations $z :=[z_1,z_2,\hdots,z_T] \in
\bO^T$ such that $\sP(Z=z)>0$. Denote
\[
\pi^{(z)} := \left[\pi_1^{(z)},\hdots,
  \pi_{T}^{(z)}\right]\quad \text{where}\quad  \pi_t^{(z)}(x):=\sP(X_t = x \mid
Z_1=z_1,\hdots,Z_t=z_t),\quad x\in\bS, \quad 1\leq t\leq T.
\]
A mathematical formalization of the objective is as follows (see
Fig.~\ref{fig:transformer_objective}) :
\begin{objective}\label{obj:goal}
Define a \textbf{fixed-point mapping} $\clN: \clP(\bS)^T \mapsto
\clP(\bS)^T$ such that 
\[
\clN \pi^{(z)} = \pi^{(z)}.
\]
\end{objective}

\section{Fixed-point inference architecture for an HMM}\label{sec:HMM}

\begin{figure}[h!]
\centering








\begin{tikzpicture}[
    node distance=1.2cm,
    latent/.style={draw, circle, minimum size=0.8cm, inner sep=0pt},
    obs/.style={draw, rectangle, minimum size=0.8cm, inner sep=0pt},
    arrow/.style={-stealth, thick},
    dots/.style={inner sep=3pt, minimum size=0.2cm}
]

    \node (start_x) {};
    \node (latent_label) [left=0.3cm of start_x] {state (hidden)};
    \node (observation_label) [above=0.7 cm of latent_label] {observation};

    \node (x0) at ($(start_x) + (0.5cm, 0)$) [latent] {$X_0$};
    \node (x1) [latent, right=of x0] {$X_1$};
    \node (dots) [dots, right=of x1] {$\cdots$};
    \node (xT_1) [latent, right=of dots] {$X_{T-1}$};
    \node (xT) [latent, right=of xT_1] {$X_T$};

    \node (z1) [obs, above right=0.5cm and 1.27cm of x0] {$Z_1$};
    \node (z2) [obs, above right=0.5cm and 1.27cm of x1] {$Z_2$};
    \node (zd) [above right=0.5cm and 1.27cm of dots] {$\cdots$};
    \node (zT) [obs, above right=0.5cm and 1.27cm of xT_1] {$Z_T$};
    \node (zT_plus_1) [obs, above right=0.5cm and 1.27cm of xT] {$Z_{T+1}$};

    \draw [arrow] (x0) -- (x1);
    \draw [arrow] (x1) -- (dots);
    \draw [arrow] (dots) -- (xT_1);
    \draw [arrow] (xT_1) -- (xT);

    \draw [arrow] (x0) -- (z1);
    \draw [arrow] (x1) -- (z2);
    \draw [arrow] (xT_1) -- (zT);
    \draw [arrow] (xT) -- (zT_plus_1);

    \draw [arrow] ($(x0) + (-0.5cm, -1.0cm)$) -- ($(xT) + (0.5cm, -1.0cm)$) node[right] {time};

    \draw ($(x0) + (0, -1.0cm)$) -- ($(x0) + (0, -1.2cm)$) node[below] {$0$};
    \draw ($(x1) + (0, -1.0cm)$) -- ($(x1) + (0, -1.2cm)$) node[below] {$1$};
    \draw ($(xT_1) + (0, -1.0cm)$) -- ($(xT_1) + (0, -1.2cm)$) node[below] {$T-1$};
    \draw ($(xT) + (0, -1.0cm)$) -- ($(xT) + (0, -1.2cm)$) node[below] {$T$};

\end{tikzpicture}
\caption{The graphical model for $(X,Z)=\text{HMM}(\mu,A,C)$.}
\label{fig:hmm}
\end{figure}
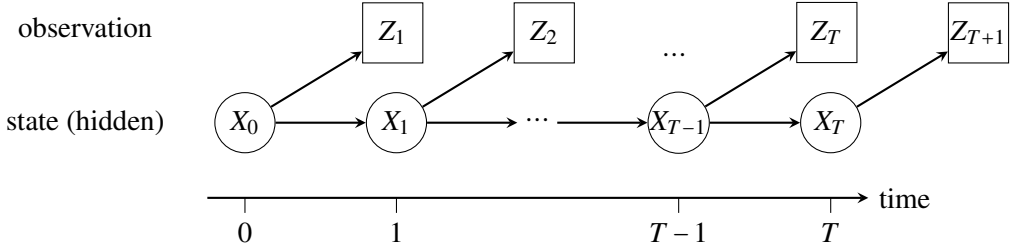

In this section, we describe the explicit form of $\clN$ for the
special case when the model is an HMM. The notation for the model is
introduced next:

\paragraph{Mathematical model.}~Consider a finite time-horizon $\mathbb{T}=\{0,1,2,\hdots,T\}$ with
$T<\infty$. An HMM is specified by a pair
of discrete-time stochastic processes $(X,Z)$ defined as follows:
\begin{enumerate}
  \item The state-space $\bS=\{1,2,\hdots,d\}$ and observation
    space $\bO=\{0,1,2,\hdots,m\}$ are finite.
\item The state process $X=\{X_t:t\in\mathbb{T}\}=\text{Markov}(\mu,A)$ is a Markov chain
  taking values in $\bS$.
\item The observation process $Z=\{Z_1,Z_2,\hdots,Z_T,Z_{T+1}\}$ takes values
  in $\bO$ and is modeled by \[
  \sP(Z_{t+1} = z \mid X_t=x) = C(x,z),\quad z\in\bO,\quad x\in\bS, \quad
  t\in\mathbb{T}.
  \]
\end{enumerate}
The model is denoted by $(X,Z)=\text{HMM}(\mu,A,C)$ (see Fig.~\ref{fig:hmm}). 

For the HMM, our goal is to describe an explicit form of the mapping $\clN$.   The algorithm in the following subsection is based on the theory for
nonlinear predictor developed in our prior
work~\citep{chang2025dual}.  A summary of the same is given in Appendix~\ref{sec:theory_HMM}.

\subsection{Fixed-point algorithm for inference}\label{sec:algo_HMM}

Let $\rho
=[\rho_1,\rho_2,\hdots,\rho_T]\in \clP(\bS)^T$ and
$z=(z_1,z_2,\hdots,z_T)\in \bO^T$.  Our goal is to define a mapping ${\cal
  N}:\rho \mapsto   \rho^+$ such that $\pi^{(z)}$ is its
fixed-point. Towards this goal, 
\begin{enumerate}
\item Define
  \[
 c_t(x):=2 \, C(x, z_{t}) - 1,\quad x\in\bS, \quad 1\leq t\leq T.
\]
\item For $\nu\in\clP(\bS)$, $f,c\in\Re^d$, define
  \[
\phi(f;\nu,c) =           \begin{cases}
          \frac{-1}{(1-\nu(c)^2)} \left( \nu((Af)(c-\nu(c))) \right), & 1-\nu(c)^2 \neq 0, \\
          0 & \text{o.w.}
         \end{cases} 
   \]
\end{enumerate}

Based on these, consider a backward difference equation (B$\Delta$E) as
follows:
\begin{subequations}\label{eq:opt_BDE}
\begin{align}
y_s(x) &= (Ay_{s+1})(x) + c_{s+1} (x) u_s,\quad  x\in\bS,\quad
         s=0,1,2,\hdots,t-1, \label{eq:opt_BDE_a} \\
u_s & =  \begin{cases} \phi(y_{s+1};\rho_s,c_{s+1}) ,&
  s=1,2,\hdots,t-1, \\
  \phi(y_{1};\mu,c_{1}), & t=0, 
\end{cases}
  \label{eq:opt_BDE_a2}
  \\
y_t(x)  & = f(x),\quad x\in\bS. \label{eq:opt_BDE_b}
\end{align}
The solution $y:=\{y_s(x)\in\Re:x\in\bS,\;0\leq s\leq t-1\}$ of the
B$\Delta$E is used to define a mapping $\clN:\clD \subset
\clP(\bS)^T\to \clP(\bS)^T$ as follows:
\begin{align}
(\clN \rho)_t := \rho_t^+\quad \text{s.t.} \quad \rho_t^+ (f) := \mu (y_0) - \sum_{s=0}^{t-1} u_s,\quad 1\leq t\leq T. \label{eq:opt_BDE_c}
\end{align}
\end{subequations}
(compare with~\eqref{eq:rep_attention_simplified}). 
The domain $\clD$ is defined to be the largest subset in $\clP(\bS)^T$ such
that $\clN \rho \in \clP(\bS)^T$.  That the domain is non-empty is because of the
following proposition which also shows the significance of~\eqref{eq:opt_BDE} to the
computation of $\pi^{(z)}$.

\begin{proposition}[Prop.~21 in \cite{chang2025dual}]\label{prop:dual_filter}
  Suppose $(X,Z)=\text{HMM}(\mu,A,C)$.  
Consider~\eqref{eq:opt_BDE}.  Then 
  \begin{equation}\label{eq:FPE}
  \clN \pi^{(z)} = \pi^{(z)}.
  \end{equation}
\end{proposition}



\section{Directions for future work}\label{sec:conc}


In this paper, inspired by the decoder-only transformer, we introduced a
probabilistic model together with an inference architecture, formalized
as Objective~\ref{obj:goal}. The architecture was described explicitly
for the special case of a hidden Markov model (HMM), which serves as a
useful starting point. A natural next step is to extend the framework
to more general classes of hidden processes $X$. In this regard, we
note: 
\begin{enumerate}
\item The positional encoding embodies a counterintuitive 
  yet apparently effective approach to
  representing time in the causal processing of sequential data.
\item The attention operation~\eqref{eq:transf_output} exhibits a symmetry:  
  permuting the inputs $\{\sigma_1,\sigma_2,\hdots,\sigma_{t-1}\}$
  (while keeping $\sigma_t$ fixed) yields the identical output $o_t^h$
  at time $t$.
\end{enumerate}
An understanding of these features, and their incorporation into
probabilistic modeling, remains an important direction for future work.

\section{Appendix}
\label{appdx}

\subsection{ChatGPT's answer to the prompt: what do you think is the intended meaning of these two questions?}
\label{appdx:chatgpt_Wolfram}

\begin{quote}
{\em This title poses two layered questions, both inviting deeper reflection on the nature of large language models.

\paragraph{1. What Is ChatGPT Doing?}
This question asks us to examine the behavior of ChatGPT at multiple levels:
\begin{itemize}
  \item \textbf{Functionally:} What kind of tasks is ChatGPT performing---answering questions, generating coherent text, simulating dialogue?
  \item \textbf{Mechanistically:} What computations or internal representations underlie these responses?
  \item \textbf{Linguistically or cognitively:} Does it understand? Is it reasoning? Or is it merely imitating human-like language?
\end{itemize}
At its core, the question invites us to describe what is happening when we observe ChatGPT in action.

\paragraph{2. Why Does It Work?}
This second question seeks a deeper explanation:
\begin{itemize}
  \item \textbf{Technically:} Why do language models trained with simple objectives (e.g., next-token prediction) generalize so well?
  \item \textbf{Conceptually:} How can models built on pattern matching exhibit coherent reasoning and creativity?
  \item \textbf{Philosophically:} What does this success suggest about the nature of intelligence, understanding, and language?
\end{itemize}
This question acknowledges a puzzle: How does a model trained on statistical correlations manage to generate responses that appear meaningful and intelligent?

Together, these two questions capture a dual aim: to observe and to explain---to understand not just what language models do, but why they are so effective in doing it.}
\end{quote}

\subsection{Proof of well-posedness}
\label{appdx:existence}

The existence theorems rely on the following proposition from
linear algebra.

\begin{proposition}\label{prop:linear_algebra}
  Let $s:\bO\to \Re$. Then there exists unique $(s,\tilde{s}) \in \Re
  \times \Re^m$ such that the following decomposition holds:
  \[
  s(z) = \bar{s} +\tilde{s}^\tp e(z) ,\quad z\in\bO.
\]
Explicitly, 
\[
\bar{s}:= \frac{1}{m+1} \sum_{z\in\bO} s(z),\quad \text{and} \quad 
\tilde{s} (i) = (s(i) - \bar{s}),\quad
i=1,2,\hdots,m.
\]
\end{proposition}
\begin{proof}
  We have 
  \[
\left( \bar{s} +\tilde{s}^\tp e(z) \right)   = \begin{cases}
\left( \bar{s} + ( s(z) -\bar{s} )\right) = s(z) , &  z=1,2,\hdots,m,\\[5pt]
   \left( \bar{s} - \tilde{s}^\tp \ones \right) = s(0),
     & z=0,
     \end{cases}
 \]
 where the last step follows because
 \[
   \tilde{s}^\tp \ones = \sum_{i=1}^m \tilde{s}(i)  = \sum_{i=1}^m
     \left(s(i) - \bar{s} \right) = - m \bar{s}  + \sum_{i=1}^m s(i),\]
and therefore,  \[
\bar{s} - \tilde{s}^\tp \ones = \bar{s} - \left(- m \bar{s} +
  \sum_{i=1}^m s(i) \right) = (m+1) \bar{s} - \sum_{i=1}^m s(i) = s(0).
\]
The decomposition is unique because $\bar{s} + \tilde{s}^\tp e(z)
\equiv 0$ implies
\begin{align*}
  \bar{s} + \tilde{s}^\tp e(i) = \bar{s} + \tilde{s}(i) &= 0, \quad i=1,2,\hdots,m, \\
  \bar{s} + \tilde{s}^\tp e(0) = \bar{s} - \sum_{i=1}^m \tilde{s}(i) &= 0.
\end{align*}
Summing the first of these equations over $i$,
\[
m\bar{s} + \sum_{i=1}^m \tilde{s}(i) = (m+1) \bar{s} =0,
\]
which then also implies $\tilde{s}(i) = -\bar{s} =0$ for
$i=1,2,\hdots,m$.         
\end{proof}

\begin{example}[m=1] Let $s:\{0,1\} \to \Re$.  Denote
$
   s^+ := s(1)$ and $s^{-} := s(0)
 $. Then
 \[
 s(z) = \bar{s} +  \tilde{s} \, e(z),\quad z\in\{0,1\},
\]
where $\bar{s}:=\half (s^+ + s^-)$ and $\tilde{s}:= \half (s^+ -
  s^-)$ (recall $e(1)=1$ and $e(0)=-1$).
\end{example}

\begin{remark}
There is nothing special about the choice of
$\{e(1),e(2),\hdots,e(m)\}$, chosen in this paper to be the canonical basis.  One could
instead choose these vectors to be any basis of $\Re^m$, and 
set $e(0)=-e(1)-e(2)-\hdots-e(m)$ as before.
In~\cite{fukasawa2023backward}, such a structure is referred to as a 
lattice. See~\citep{cohen2010general} for a 
general theory of BS$\Delta$E.
\end{remark}

We now provide a proof
  of~\Prop{prop:existence_nonlin_predictor_rep} leading to well-posedness of
  representation~\eqref{eq:nonlin_predictor_rep}.

\begin{proof}[of \Prop{prop:existence_nonlin_predictor_rep}]
We begin by proving a result where the representation is shown to hold
for {\em any} $S_T \in \clZ_T$.  From Doob-Dynkin lemma, there is a
deterministic function $s:\bO^T \to \Re$ such that
\[
S_T = s(Z_1,\hdots,Z_{T-1},Z_T).
\]
Set
\[
S(z) := s(Z_1,\hdots,Z_{T-1},z),\quad z\in\bO.
\]
From~\Prop{prop:linear_algebra},
\[
S_T = S(Z_T) = S_{T-1} - (U_{T-1})^\tp e(Z_T),
\]
where
\[
  S_{T-1} = \frac{1}{m+1} \sum_{z\in \bO} S(z),\quad 
 U_{T-1}(i):= - (S(i) - S_{T-1}),\quad
i=1,2,\hdots,m.
\]
Uniqueness is from the uniqueness of the decomposition. The proof is completed through induction by
repeating the procedure for $S_{T-1}\in\clZ_{T-1}$.

A direct application of the above result to justify the
representation~\eqref{eq:nonlin_predictor_rep} for conditional expectation is
complicated by a subtle issue: The conditional expectation
is meaningfully defined only for sample paths $Z=z$ with 
$\sP([Z=z])>0$.  Note here that because $|\bO|=m+1$ and $T$ are
both finite, there are only finitely many---specifically $(m+1)^T$---sample paths.
Thus, $\sP([Z=z])$ is a well-defined object for each sample path, 
although it may be zero depending on the model properties.

There are two ways to address this issue: 
\begin{enumerate}
\item Assume 
  $\sP([Z=z])\geq\underline{c}^T > 0$ for all $z\in\bO^T$,
and the existence of a unique $U$ follows directly from the earlier result.
\item Adopt the convention $\frac{0}{0}=0$ to define (or extend) the
  conditional expectation for sample 
paths $Z=z$ with $\sP([Z=z])=0$. Then again, a particular
selection of $U$ follows from the above result.
\end{enumerate}
In the second case, however, there
may be other choices of $U$ such that the
representation~\eqref{eq:nonlin_predictor_rep} holds: Any two choices
will yield a representation that coincides on the set $\{z\in \bO^{T}:
\sP(Z=z)>0\}$ but may differ on the set $\{z\in \bO^{T}:
\sP(Z=z)=0\}$.
\end{proof}

\begin{example}[m=1] 
Set $
S_T^{+} = s(Z_1,\hdots,Z_{T-1},1)$ and $S_T^{-} =
s(Z_1,\hdots,Z_{T-1},0)$. 
Then $S_T^{+}, S_T^{-}\in \clZ_{T-1}$ and
\[
S_T = S_{T-1} - U_{T-1} e(Z_T),
\]
where $
S_{T-1}=\half\left( S_T^{+} + S_{T}^-\right)$ and $U_{T-1}= -\half\left( S_T^{+} - S_{T}^-\right)$. 
\end{example}

\subsection{Additional operations in a transformer}
\label{sec:xfer}

  This section is based on~\citep[Ch.,~10]{jm3}.  Fix $t$. The output $y_t$ from concatenating the output of multiple heads is subject to the following operations:
  \begin{enumerate}
  \item Residual connection which means
    \[
 y_t\mapsto y_t + e_t.
    \]
  \item Layer normalization which means
    \[
  y_t\mapsto \text{diag}(\gamma) \frac{y_t-\text{mean}(y_t)}{\text{std}(y_t)} + \beta.
  \]
  where $\gamma\in\Re^d$ and $\beta\in\Re^d$ are learnable parameters (referred to as gain and offset).
\item Feedforward neural network
  \[
  y_t\mapsto \text{FFN}(y_t).
  \]
\end{enumerate}

\newP{Summary of operations in a single layer.}
 Fix time $t$.  The following operations define a single layer in a transformer:
 \begin{align*}
   y_t & = \text{MultiHeadAttention}(e_t,[e_1,e_2,\hdots,e_{t-1}])\\
   y_t & = y_t + e_t \\
   y_t &= \text{LayerNorm}(y_t) \\
   y_t & = y_t + \text{FFN}(y_t) \\
   y_t & = \text{LayerNorm}(y_t)
 \end{align*}
 The learnable parameters in the MultiHeadAttention are the matrices $W_V,W_K,W_Q,W_O$.  For each of the two LayerNorm operations, the learnable parameters are the gains $\gamma$ and the offset $\beta$.  Additionally, the weights of the FFN are also learned.

\subsection{Theory for the nonlinear predictor} \label{sec:theory_HMM}

Recall the
  representation~\eqref{eq:nonlin_predictor_rep}.  Our goal is to
  provide an interpretation of $U$ as an optimal control input for a
  certain (dual) optimal control problem\footnote{The result is a
    generalization of the theory for linear
    predictor~\eqref{eq:linear_predictor} for Gauss-Markov process.
    For such a process, $u$ in the linear
    predictor~\eqref{eq:linear_predictor} is an optimal control input
    for a deterministic 
    linear quadratic (LQ) optimal control problem.}.  The optimal
  control interpretation yields a formula for $U$, from which an
  explicit form of the mapping $\clN$ (described in \Sec{sec:algo_HMM}) is derived.

The optimal control formulation is based on some recent advances in duality theory~\citep{kim2025arrow}: 
An HMM is an example of a state-output system ($X$ is the state and
$Z$ is the output). In analogy with the 
classical duality in linear systems theory, we define an input-state system that is dual to an HMM. To this end, define
\[
c(x) := \begin{bmatrix} C(x,1) - C(x,0) \\ C(x,2) - C(x,0) \\ \vdots \\
  C(x,m) - C(x,0) \end{bmatrix}_{m\times 1},\quad x\in\bS.
\]
For each fixed $x\in \bS$, $c(x)$ is a $m\times 1$ vector.

\newP{Dual control system}~is a backward stochastic difference equation 
(BS$\Delta$E) 
\begin{subequations}\label{eq:dual_BSDE}
\begin{align}
Y_t(x) &= (AY_{t+1})(x) + c^\transpose(x) (U_t + V_t(x)) - V_t^\transpose(x)
         e(Z_{t+1}),
  \quad \;\; x\in\bS,\quad t=0,1,2,\hdots,T-1, \label{eq:dual_BSDE_a}\\
Y_T(x)  & = F(x),\quad x\in\bS,  \label{eq:dual_BSDE_b}
\end{align}
\end{subequations}
where $U\in\clU$ is the control input and the
terminal condition $F \in \clZ_T$ (which means $F(x)$ is allowed to be
$\clZ_T$-measurable for each $x\in\bS$).  For a given $U$ and $F$, the
BS$\Delta$E~\eqref{eq:dual_BSDE} is solved to obtain the solution
$(Y,V)=\{(Y_t(x),V_t(x))\in\clZ_t :x\in\bS,\;t=0,1,2,\hdots,T-1\}$
(see~\cite[Prop.~10]{chang2025dual} where the existence and uniqueness
of the solution is shown).

Fix $U\in \clU$.  For the solution $(Y,V)$ of the
BS$\Delta$E~\eqref{eq:dual_BSDE}, define  
\begin{align*}
\sJ_T(U;F)  :=
\dvar (Y_0(X_0)) + 
\E \Big( \sum_{t=0}^{T-1} l (Y_{t+1},V_t,U_t\,;X_t)  \Big),
\end{align*}
where $\dvar(Y_0(X_0))=\E(|Y_0(X_0) - \mu(Y_0)|^2) = \mu(Y_0^2)
-\mu(Y_0)^2$ (note here $Y_0$ is a deterministic function), and the running cost $l:\Re^d\times\Re^{m\times
  d}\times\Re^m\times\bS\to\Re$ is given by,
  \[
  l(y,v,u;x):=  (\Gamma y)(x) + (u+v(x))^\transpose R(x) (u+v(x)),
\quad y\in\Re^d,\;v\in\Re^{m\times d},\;u\in\Re^m,\;x\in\bS,
\]
where
 \begin{align*}
 (\Gamma f)(x) &:= \sum_{y\in\bS} A(x,y) f^2(y) - (Af)^2(x),\quad x\in
   \bS, \\
  R(x) &:=  \text{diag}(c(x)) + C(x,0) (I+\ones \ones^\transpose)  - c(x) c^\transpose (x), \quad x\in\bS.
 \end{align*}

Now that the dual control system and the optimal control objective have
been defined, the duality principle is given in the following
theorem.

\begin{theorem}[Thm.~13 in~\cite{chang2025dual}]\label{thm:duality-principle}
	Let $U\in \clU$ and $F\in\clZ_T$.  Consider an estimator
	\begin{equation*}\label{eq:estimator}
		S_T := \mu(Y_0) - \sum_{t=0}^{T-1} U_t^\transpose e(Z_{t+1}).
	\end{equation*}
	Then 
	\begin{equation*}\label{eq:duality-principle}
		\sJ_T(U;F) = \E\big(|F(X_T)-S_T|^2\big).
	\end{equation*}	
      \end{theorem}

Noting that the right-hand side is the
mean-squared error, the duality principle provides for an optimal
control approach to
compute the conditional expectation. 

\begin{itemize}
\item \textbf{Dual optimal control problem (OCP):}
\end{itemize}
\begin{equation}\label{eq:dual-optimal-control}
    \min_{U \in \clU} \; \sJ_T(U;F) 
    \quad \text{subject to} \quad \eqref{eq:dual_BSDE}
  \end{equation}

The relationship between the OCP~\eqref{eq:dual-optimal-control} and the nonlinear filter is as
follows:
 \begin{proposition}[Prop.~14 in \cite{chang2025dual}]\label{prop:nonlinear_predictor_optimal_control}
Consider OCP~\eqref{eq:dual-optimal-control}.  For this problem, there
exists an optimal control 
  $U^\opt = \{U_t^\opt:0\le t \le T-1\} \in \clU$ such that,
\begin{align*}
		\pi_T(F) = \mu(Y_0^\opt) - \sum_{t=0}^{T-1} (U_t^\opt)^\tp
  e(Z_{t+1}),\quad \sP\text{-a.s.},\quad F\in \clZ_T,
\end{align*}
where $Y_0^\opt$ is obtained from solving the
BS$\Delta$E~\eqref{eq:dual_BSDE} with $U=U^\opt$.  The optimal value
is given by
\[
\sJ_T(U^\opt;F) = \E(|F(X_T)-\pi_T(F)|^2) = \text{MMSE}.
\]
      \end{proposition}

      \begin{remark}
Taking the function $F$ as $F(x)=C(x,z)$ yields a nonlinear predictor
for $p_T(z)$, for $z\in\bO$. 
      \end{remark}

The solution of the OCP~\eqref{eq:dual-optimal-control} leading to an 
explicit formula for the optimal control $U^\opt$ is described next.  

\newP{Formula for optimal control} Denote
\[
\phi(y,v;\rho) :=    
          - \rho(R)^{\dagger} \left( \rho ((c-\rho (c))y)- \rho (R
            v)\right), \quad y\in\Re^d, \; v\in \Re^{m\times d},\; \rho\in\clP(\bS).
       \]
        Here, $\rho(R)^{\dagger}$ denotes the pseudo-inverse of the
       $m\times m$ matrix $\rho(R):=\sum_{x\in\bS} \rho(x) R(x)$. The
       other two terms
       are $\rho ((c-\rho (c))y) := \sum_{x\in\bS} \rho(x) (c(x) -
       \rho(c)) y(x)$ and $\rho (R
            v):= \sum_{x\in\bS} \rho(x) R(x)v(x)$.  Note that both of
            these terms are $m\times 1$
            vectors.

\begin{theorem}[Thm.~17 in \cite{chang2025dual}]\label{thm:optimal-solution}
	Consider the OCP~\eqref{eq:dual-optimal-control}. Then 
	an optimal control is of the feedback form given by
        \begin{subequations}
	\begin{equation}\label{eq:optimal_control_formula}
         U_t^\opt = \phi(Y_t,V_t;\pi_t),\quad \sP\text{-a.s.},\quad 0\leq
       t\leq T-1.
       \end{equation}
	Suppose $(Y^\opt,V^\opt)$ is the
	 solution of the
	BS$\Delta$E~\eqref{eq:dual_BSDE} with $U=U^\opt$.  Then the
        following representation holds:
		\begin{align}
		\pi_t (Y_t^\opt) = \mu (Y_0^\opt) -
                   \sum_{s=0}^{t-1} (U_s^\opt)^\tp e(Z_{s+1}),\quad
                   \sP\text{-a.s.},\quad 0\leq
       t\leq T.
                   \label{eq:estimator-t}
		\end{align}
	\end{subequations}
              \end{theorem}

 \begin{remark}
Compare~\eqref{eq:estimator-t} with the representation given in
\Prop{prop:nonlinear_predictor_optimal_control}.  While the
representation in
\Prop{prop:nonlinear_predictor_optimal_control} is given only for the terminal time
$T$,~\Thm{thm:optimal-solution} shows that the optimal control input
in fact yields a representation for the entire conditional measure $\pi_t$ for $0\leq t\leq T$. 
\end{remark}

Denote
\[\Pi  :=\{\rho:\Omega\times \{1,2,\hdots,T\} \to \clP(\bS), \;\;
  \rho_t\in\clZ_t, \;1\leq t\leq T\}.
\]
Note that the conditional measure $\pi=\{\pi_t:1\leq t\leq T\}$ is an
element of $\Pi$.  The subset of deterministic
probability measures in $\Pi$ is denoted by $\Pi^\det$.

\newP{Fixed-point mapping} Pick $\rho\in \Pi$ and consider the BS$\Delta$E control system,
\begin{subequations}\label{eq:opt_BSDE}
\begin{align}
Y_t(x) &= (AY_{t+1})(x) + c^\transpose (x) (U_t +
         V_t(x)) - V_t^\transpose (x)
         e(Z_{t+1}),\quad  x\in\bS,\quad
         t=0,1,2,\hdots,T-1, \label{eq:opt_BSDE_a} \\
U_t & = \phi(Y_t,V_t;\rho_t) ,\quad
         t=0,1,2,\hdots,T-1, \label{eq:opt_BSDE_a2}
  \\
Y_T(x)  & = F(x),\quad x\in\bS.\label{eq:opt_BSDE_b}
\end{align}
Although the formula for the optimal control law is applied, the
resulting control input is generally sub-optimal, since $\rho$ may not
coincide with the true conditional
measure $\pi$. 

Because the terminal condition $Y_T=F$ is arbitrary, the solution $Y$
induces a mapping $\clN:\clD \subset \Pi \to \Pi$ as follows:
\begin{align}
(\clN\rho)_t(Y_t) := \mu (Y_0) - \sum_{s=0}^{t-1} U_s^\transpose
e(Z_{s+1}),\quad 0\leq t\leq T,\label{eq:opt_BSDE_c}
\end{align}
\end{subequations}
where on the right-hand side, $\{U_s:0\leq s\leq t-1\}$ is according to~\eqref{eq:opt_BSDE_a2}
and~$Y_0$ is the solution of~\eqref{eq:opt_BSDE_a} at
time $t=0$.   It is clear that $\clN\rho \in \clY^\dagger$ because the right-hand
side of~\eqref{eq:opt_BSDE_c} is well-defined and $\clZ_t$-measurable for each $0\leq t\leq T-1$.
The domain $\clD$ is defined to be the largest subset in $\Pi$ such
that $\clN\rho \in \Pi$.  That the domain is non-empty is because of the
following corollary of \Thm{thm:optimal-solution}.

\begin{corollary}[Corollary~20 in \cite{chang2025dual}]
  Consider~\eqref{eq:opt_BSDE}.  Then $\pi\in \clD$ and moreover
  \[
  \clN \pi = \pi,\quad \sP\text{-a.s.}
  \]
\end{corollary}
\begin{proof}
With $\rho=\pi$, because $\phi$ is
used in~\eqref{eq:opt_BSDE_a2},~\eqref{eq:opt_BSDE_c} is given by~\eqref{eq:estimator-t}. 
\end{proof}

\newpage

\bibliography{bibfiles/_master_bib_jin.bib,bibfiles/jin_papers.bib,bibfiles/extrabib.bib,bibfiles/estimator_controller.bib,bibfiles/transformer.bib,bibfiles/transformer_math.bib}

\end{document}